\documentclass[journal]{IEEEtran}
\IEEEoverridecommandlockouts
\usepackage{cite}
\usepackage{amsmath,amsthm,amssymb,amsfonts}
\usepackage{array}
\usepackage{graphicx}
\usepackage{textcomp}
\usepackage{bbold}
\usepackage[mathscr]{eucal}
\usepackage[table]{xcolor}
\usepackage{xcolor}
\usepackage{cleveref}
\crefformat{section}{\S#2#1#3}
\usepackage[framemethod=tikz]{mdframed}
\usepackage[ruled,vlined,linesnumbered]{algorithm2e}
\usepackage{algpseudocode}
\usepackage{multirow}
\usepackage{rotating}

\usepackage[font=footnotesize]{caption}
\usepackage{subcaption}
\usepackage{pifont}
\usepackage[flushmargin]{footmisc}

\algnewcommand\LeftComment[2]{%
\hspace{#1\algindent}$\triangleright$ \eqparbox{COMMENT}{#2} \hfill %
}

\let\oldnl\nl
\newcommand{\nonl}{\renewcommand{\nl}{\let\nl\oldnl}}

\def\BibTeX{{\rm B\kern-.05em{\sc i\kern-.025em b}\kern-.08em
    T\kern-.1667em\lower.7ex\hbox{E}\kern-.125emX}}
\begin{document}
\title{A Dataset of Inertial Measurement Units for Handwritten English Alphabets}
\author{\IEEEauthorblockN{Hari Prabhat Gupta, Senior Member IEEE, Rahul Mishra, Member IEEE\\Department of Computer Science and Engineering, IIT (BHU) Varanasi}
\thanks{\textit{Instructors:} Hari Prabhat Gupta and Tanima Dutta, Senior Member, IEEE\\
\textit{Report writing:} Rahul Mishra, Member, IEEE, and Garvit Banga\\
\textit{TA:} Shubham Pandey, Krishna Sharma, and  Himanshu Sahu\\
\textit{Volunteers}~\cite{vol1} 
}}

\maketitle
\begin{abstract}
This paper presents an end-to-end methodology for collecting datasets to recognize handwritten English alphabets by utilizing Inertial Measurement Units (IMUs) and leveraging the diversity present in the Indian writing style. The IMUs are utilized to capture the dynamic movement patterns associated with handwriting, enabling more accurate recognition of alphabets. The Indian context introduces various challenges due to the heterogeneity in writing styles across different regions and languages. By leveraging this diversity, the collected dataset and the collection system aim to achieve higher recognition accuracy. Some preliminary experimental results demonstrate the effectiveness of the dataset in accurately recognizing handwritten English alphabets in the Indian context. This research can be extended and contributes to the field of pattern recognition and offers valuable insights for developing improved systems for handwriting recognition, particularly in diverse linguistic and cultural contexts. 

\end{abstract}

\section{Introduction}
Recognizing English alphabets using Inertial Measurement Units is a promising field of research that opens new possibilities for the development of sign language recognition systems, virtual reality, and gesture-based interfaces~\cite{8936464,8254518}. It involves the use of IMU sensors to recognize and identify alphabets based on the movements of a person's hand. IMUs are small electronic devices that can detect and measure changes in motion, orientation, and direction using a tri-axial accelerometer, gyroscope, and magnetometer, respectively~\cite{mishra2019fault,8328265, 9768100}. Due to the low cost, low energy consumption, and compact size of the IMU sensors, they are widely used in Internet of Things (IoT) applications. Thus, creating a state-of-the-art in this field of sensors-based recognition~\cite{10041779,9653808,10.1145/3570955,8275960}. For example, using the data sign language recognition systems can benefit significantly from this IMU-based collected data, as it can improve the accuracy and speed of recognition, making it easier for people with hearing impairments to communicate. Similarly, in virtual reality and gesture-based interfaces~\cite{8328238}, IMUs’ data can provide a more immersive and interactive experience, allowing users to control and manipulate objects in a more natural and intuitive way~\cite{8328265,9852654}.

While English alphabet recognition using IMUs has shown a lot of promise, there are still challenges and limitations that need to be addressed to make this technology more effective and accessible. For instance, the accuracy of recognition can be affected by factors such as lighting conditions, the size of the dataset, and the complexity of the gestures~\cite{mishra2022designing,7498690, mishra2022suppressing}. Moreover, the cost of the IMU hardware can be a barrier to adoption, particularly in low-income regions. This dataset collection aims to provide vast and heterogeneous data of English alphabets (both upper and lower cases) from more than 100 users. It will provide a deeper insight into the principles and techniques behind English Alphabets Recognition using IMUs. It will provide an overview of the current state-of-the-art in this field, including the various approaches and algorithms that are used to recognize alphabets based on motion data. Additionally, this description of the dataset will discuss the potential applications of this technology in various domains and industries. Our dataset is available at IEEE Dataport~\cite{av6q-jj17-23} and Watch the data collections treasure on youtube~\cite{you1}.

\subsection{Aim of dataset collection}
The principle aim of this dataset collection is summarized in the following points:
\begin{itemize}
\item To develop a large alphabets dataset while considering a large number of heterogeneity parameters, including participants' height, gender, days, and timing of data collection.
\item To facilitate an abundant sensory dataset for recent trending technologies such as Federated Learning for performing real-world evaluation alongside the simulation on machine-generated datasets or homogeneous datasets.
\item To strengthen the development of more robust and versatile datasets for validating the newly generated security and performance policies in the smart sensing environment. 
\end{itemize}

\subsection{Target applications}
\begin{itemize}
\item[]\textbf{Natural language processing (NLP):} This alphabet recognition dataset can be used as a starting point for building NLP models, which can analyze and understand human language. NLP models are used in applications such as chatbots, language translation, and sentiment analysis. 
\item[]\textbf{Fraud detection:} This dataset can be used to detect fraudulent documents or transactions, such as forged signatures or altered checks.
\item[]\textbf{Handwriting analysis:} This dataset can be used to analyze and identify handwriting styles, which can be useful in forensic investigations or in analyzing historical documents. 
\item[]\textbf{Education:} The character recognition dataset is used in educational applications to help students learn to read and write, and to develop language skills. For example, handwriting recognition software can be used to help students practice writing letters and words. 
\item[]\textbf{Accessibility:} The character recognition dataset is used to develop assistive technologies for people with disabilities. For example, OCR technology can be used to convert printed text into speech for people who are blind or have low vision.
\end{itemize}
    
\textbf{Paper Organization:} Section~\ref{device} provides an overview of the data collection devices. Section~\ref{setup} outlines the detailed information about the setup for data collection. Section~\ref{vol} and Section~\ref{des}  cover volunteers' details and data description, respectively. Finally, Section~\ref{conc} concludes this work.

\section{Data collection devices }\label{device}
In order to optimize the process of collecting sensory data from the marker pen, it is crucial to address several key questions before developing the necessary data collection devices. These questions are essential for ensuring both the effectiveness and efficiency of the data collection process. By thoroughly considering these aspects, we can create an optimal marker pen with sensing capabilities, which meets the desired objectives. 
The key question while making data collection devices are as follows:

\begin{itemize}

\item[\textbf{Q1:}] Where can we place the sensors on the marker pen? Specifically, determining the optimal location for sensor integration on the marker pen, including the top, middle and bottom.

\item[\textbf{Q2:}] What is the impact of placing sensors in different locations on the recognition performance?

\item[\textbf{Q3:}] Weather sensors' placement location change with volunteers' height, age, gender, or placement of whiteboards?

\item[\textbf{Q4:}] Is there any relation between the inclination of the whiteboard and volunteers' ease of writing in correspondence with the sensors' placement on the marker pen?

\item[\textbf{Q5:}] Which is the most suitable sampling rate for collecting data from the marker pen?

\item[\textbf{Q6:}] What factors are to be considered while choosing the microcontroller unit for transmitting and managing the data collected from the sensors?

\item[\textbf{Q7:}] What are the challenges that may be encountered while using battery power on the marker pen? What are the measures taken into consideration to resolve the battery issues?

\item[\textbf{Q8:}] What are the possibilities of providing a power supply via a wired connection? Which types of wire the power supply is good at? 

\item[\textbf{Q9:}] Which mode of data transmission is chosen: wired or wireless?
 
\item[\textbf{Q10:}] Which wireless mode of transmission is the best fit in the context of data collection from the sensors? Both in terms of power and data rate.

\item[\textbf{Q11:}] What is the frequency of recharging the battery? If the marker pen is operating on the battery power.

\item[\textbf{Q12:}] What is the preferable (or suitable) height for placing the whiteboard?

\end{itemize}

In order to circumvent the above challenges, we employed three distinct configurations of markers with IMU sensors so that the dataset would be more broadly applicable rather than sensor-specific. Two Arduino Nano 33 BLE senses equipped with LSM9DS1, a 9-axis inertial sensor were utilized. For the Arduino Nano BLE, we employed two different configurations, one near the marker's tip and the other at the upper end. The specification of LSM9DS1 is shown in Table 1.

 \begin{figure}[h]
 \centering
\includegraphics[scale=0.36]{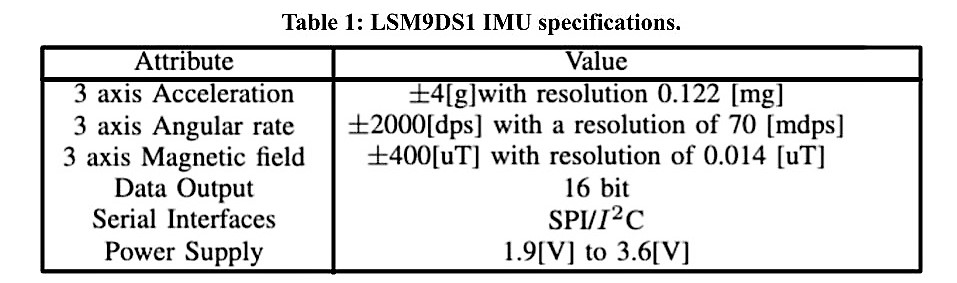} 
\captionsetup{labelformat=empty}
 \end{figure}
 
\begin{figure}[h]
 \centering
\includegraphics[scale=0.36]{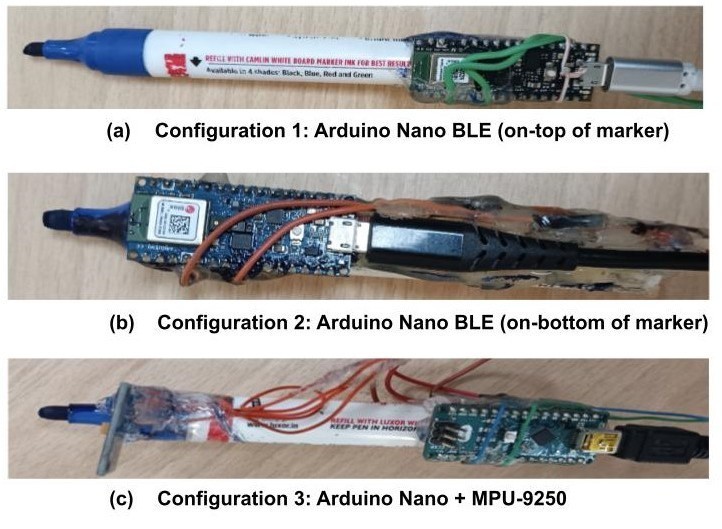} 
\caption{Illustration of different configurations of the indigenously constructed data collection devices.}
\label{fig1}
 \end{figure}

\begin{figure}[h]
 \centering
\includegraphics[scale=0.36]{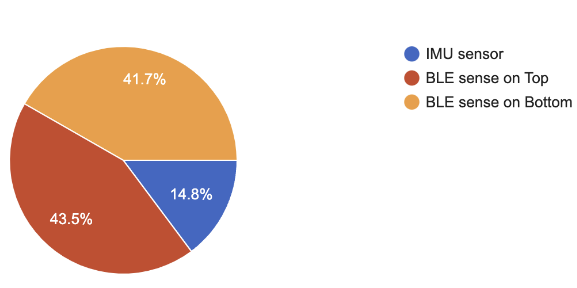} 
\caption{Illustration of the distribution of volunteers in different configurations of the indigenously constructed data collection devices.
}
\label{fig2}
 \end{figure}

In order to create the third marker, we linked an Arduino Nano to the MPU9250 9-Axis Attitude Gyro Accelerator Magnetometer Sensor Module, as depicted in Fig.~\ref{fig1}. Additionally, the distribution of volunteers using these devices is depicted in Fig.~\ref{fig2}

The details of the micro-controller and the sensors used while constructing the data collection device are as follows:

\subsection{Arduino Nano 33 BLE Sense}
It is a compact development board based on the popular Arduino platform~\cite{ble}. It is designed specifically for projects requiring Bluetooth Low Energy (BLE) connectivity and a wide range of sensors. The board is an improved version of the Arduino Nano, offering additional features and capabilities. The key components of the Arduino Nano 33 BLE Sense are as follows:

\begin{itemize}

\item[\textbf{a).}] \textit{Microcontroller:} The board is powered by a 32-bit ARM Cortex-M4 processor running at 64 MHz. It provides a good balance between performance and power efficiency.

\item[\textbf{b).}] \textit{Bluetooth Connectivity:} The Nano 33 BLE Sense has built-in Bluetooth 5.0, allowing it to communicate wirelessly with other BLE-enabled devices such as smartphones, tablets, and other Arduino boards.

\item[\textbf{c).}] \textit{Sensors:} One of the distinguishing features of the Nano 33 BLE Sense is its wide range of onboard sensors. These sensors include a 9-axis IMU (accelerometer, gyroscope, magnetometer), pressure sensor, humidity and temperature sensor, proximity sensor, and microphone. These sensors enable the board to gather data from its environment and make it suitable for various IoT and wearable projects.

\item[\textbf{d).}] \textit{GPIO Pins:} The board offers a total of 22 digital input/output (I/O) pins, among which 14 can be used as pulse width modulation (PWM) outputs. It also provides 6 analogue input pins. 

\item[\textbf{e).}] \textit{Memory and Storage:} The Nano 33 BLE Sense has 256 KB of flash memory for storing program code and 32 KB of SRAM for variable storage. Additionally, it has an onboard QSPI Flash chip that provides 2 MB of additional storage space for data and files. 

\item[\textbf{f).}] \textit{USB Interface:} The board can be powered and programmed via a micro USB connector, making it easy to connect to a computer for development purposes.

\item[\textbf{g).}] \textit{Compatibility:} The Arduino Nano 33 BLE Sense is fully compatible with the Arduino ecosystem. It can be programmed using the Arduino IDE, which supports a wide range of libraries and examples, simplifying the development process.
\end{itemize}

\subsection{Arduino Nano}
It is a compact and versatile microcontroller board that is part of the Arduino family of open-source hardware and software~\cite{nano}. It is designed for small-scale projects that require a low-cost, low-power microcontroller with a small form factor. The key features and characteristics of the Arduino Nano: 

\begin{itemize}
\item[\textbf{a).}] \textit{Microcontroller:} Arduino Nano is based on the Atmel ATmega328P microcontroller. It operates at a clock speed of 16 MHz and has 32KB of flash memory for storing program code, 2KB of SRAM for data storage, and 1KB of EEPROM for non-volatile storage.

\item[\textbf{b).}] \textit{Size and Form Factor:} The board has a small form factor, measuring only 45mm x 18mm, making it suitable for space constraints or when a small size is desired. 

\item[\textbf{c).}] \textit{Connectivity:} The Nano board comes with a USB interface, allowing it to be easily connected to a computer for programming and power supply. It also has 14 digital input/output pins, 8 analogue inputs, and 6 pulse-width modulation (PWM) outputs, providing flexibility for various applications. 

\item[\textbf{d).}] \textit{Power Options:} The Nano can be powered through a USB connection or an external power source, such as a battery. It has a built-in voltage regulator that can handle input voltages ranging from 7V to 12V, making it compatible with a wide range of power.. 

\item[\textbf{e).}] \textit{Programming:} Arduino Nano can be programmed using the Arduino Software (IDE), which is a user-friendly programming environment that allows you to write, compile, and upload code to the board. It supports the Arduino language, which is based on C/C++. 

\item[\textbf{f).}] \textit{Shields and Accessories:} The Nano board is compatible with a variety of Arduino shields and accessories, which are add-on modules that extend its capabilities. These shields can provide additional features like WiFi, Bluetooth, motor control, and display interfaces.

\item[\textbf{g).}] \textit{Open-Source:} Like other Arduino boards, the Nano is open-source, which means that its design files, schematics, and source code are freely available. This enables the community to contribute improvements, and modifications, and create derivative designs.

\end{itemize}

\subsection{IMU MPU-9250}
It is a compact and versatile integrated circuit that combines multiple sensors to provide accurate motion tracking and orientation estimation~\cite{mpu}. IMU stands for Inertial Measurement Unit, which refers to a device that measures and reports motion-related parameters such as acceleration, angular velocity, and magnetic field strength. The MPU-9250 is specifically designed for applications that require precise motion sensing, such as robotics, drones, virtual reality systems, and wearable devices. It is developed by InvenSense, a leading manufacturer of MEMS (Micro-Electro-Mechanical Systems) sensors.

The MPU-9250 integrates several sensor components into a single chip, including a 3-axis accelerometer, a 3-axis gyroscope, and a 3-axis magnetometer. Each of these sensors provides important data about the device's motion and orientation in three dimensions. The accelerometer measures linear acceleration, allowing the detection of changes in velocity and the determination of the device's tilt or inclination. This information is crucial for applications that require movement tracking, stabilization, or gesture recognition. The gyroscope measures angular velocity, providing data about rotational movements and changes in orientation. 

The magnetometer measures the strength and direction of the magnetic field around the device. It can be used to determine the device's heading or to compensate for magnetic disturbances that may affect the accuracy of the accelerometer and gyroscope readings. In addition to the sensor components, the MPU-9250 includes an on-chip digital motion processor (DMP) that offloads complex motion processing tasks from the main microcontroller or processor. The DMP can perform sensor fusion algorithms to combine data from multiple sensors, providing a more accurate and stable estimation of the device's motion and orientation. The MPU-9250 communicates with an external microcontroller or processor through an I2C (Inter-Integrated Circuit) or SPI (Serial Peripheral Interface) interface, allowing for easy integration into various electronic systems. The specification of the indigenous constructed devices for the data collection is illustrated in Table~\ref{tab2}.

\begin{table}[h]
\renewcommand\thetable{2}
\centering
\caption{Specifications of the indigenously constructed data collection devices.}
\begin{tabular}{|c|c|p{4.0cm}|}
\hline
\textbf{S.No.}  &   \textbf{Criteria}     &   \textbf{Description} \\ \hline
1.              &    Sensors placement    & We placed the sensors both on the top and bottom of the marker. \\ \hline
2.              &    Impact of placement  & Varies with the height and character to be written. \\ \hline
3.              &    Sampling rate        & We set the sampling rate to 50 Hz.\\ \hline
4.              &  Micro-controller units & We consider Arduino Nano BLE sense and Arduino Nano. \\ \hline
5.              &  Power source           & Wired, can be wireless (not applicable in this data collection). \\ \hline
6.              &  Communication          & Wired (minimizes delay in real-time). \\ \hline

\end{tabular}
\label{tab2}
\end{table}

\subsection{Experimental Evidence 1}
\begin{figure*}[h]
 \centering
\includegraphics[scale=0.52]{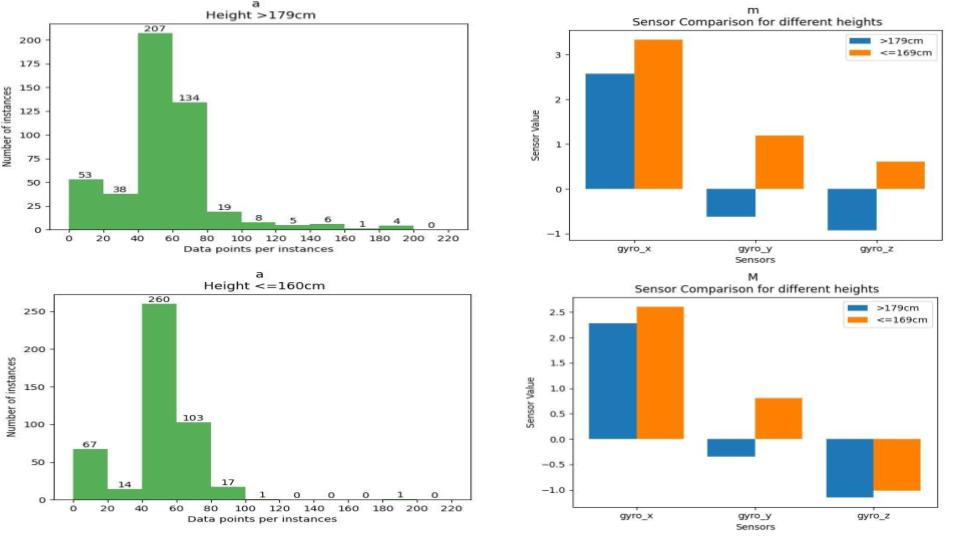} 
\caption{Impact of height on the number of data instances and values of the gyroscope sensor.}
\label{fig3}
 \end{figure*}

We present some experimental evidence, which is gathered through controlled experiments or observations to test the collected dataset.  Fig.~\ref{fig3} provides a visual representation, depicting the impact of volunteers' height on the number of data instances and values of the gyroscope sensors.

 \section{Data collection setup}\label{setup}
 Upon addressing the challenges of data collection devices, the subsequent challenges lie in establishing an appropriate data collection setup. In particular, this pertains to determining the placement of the whiteboard within our data collection setup. Similar to the challenges encountered with the data collection devices, there are different challenges involved in creating a suitable data collection setup. Such challenges can be encapsulated in the form of the following set of questions, which play a crucial role in ensuring the effectiveness and efficiency of the data collection process. Thoroughly considering these aspects allows us to devise an optimal data collection setup that aligns with our desired objectives. The fundamental questions to consider when constructing the data collection setup include:

\begin{itemize}

\item[\textbf{Q1:}] Where can we place the whiteboard? Specifically, determining the optimal location for whiteboard placement, including on the table, hanging on the wall, and on the floor.

\item[\textbf{Q2:}] What is the impact of placing whiteboard on different locations on the recognition performance?

\item[\textbf{Q3:}] Weather whiteboard location changes with volunteers height, age, gender, or style of writing?

\item[\textbf{Q4:}] Is there any relation between inclination of white board and volunteers ease of writing on the whiteboard?

\item[\textbf{Q5:}] Which is the most suitable inclination angle while placing the white board on the table? 

\item[\textbf{Q6:}] What  external factors (lightsouce, easier accessibility, air cooling vent or heating vent, isolated availability of the palce)   are to be considered while choosing the location for placement of the whiteboards?

\item[\textbf{Q7:}] What is the appropriate number of white boards to be used while data collcetion?

\item[\textbf{Q8:}] Is there any relationship between number of whiteboard counts and number of volunteers ?

\item[\textbf{Q9:}] Is the external noise hamper the performance of data collection? Is is required to place the board in noise isolated environment?

\item[\textbf{Q10:}] How to handle the interaction between the marker pen and the whiteboard surface that can introduce variability in the data? Factors such as the smoothness or texture of the whiteboard, pressure applied by the user, and the angle of the pen can affect the data collection.

\item[\textbf{Q11:}] How to monitor pressure applied by the different volunteers? Is the variation in pressure also impact the quality of the collected dataset from the sensor module?

\item[\textbf{Q12:}] How to manage angle of inclination between pen and the whiteboard? How to set whiteboard on an appropriate angle to perform optimal data collection?

\item[\textbf{Q13:}] Volunteers opinion are mandatory for effective data collection. How to provide mechanism of feedback from the volunteers to improve the data collection setup?

\item[\textbf{Q14:}]  How to manage angle of inclination between pen and the whiteboard? How to set whiteboard on an appropriate angle to perform optimal data collection?

\end{itemize}

\begin{figure}[h]
 \centering
\includegraphics[scale=0.22]{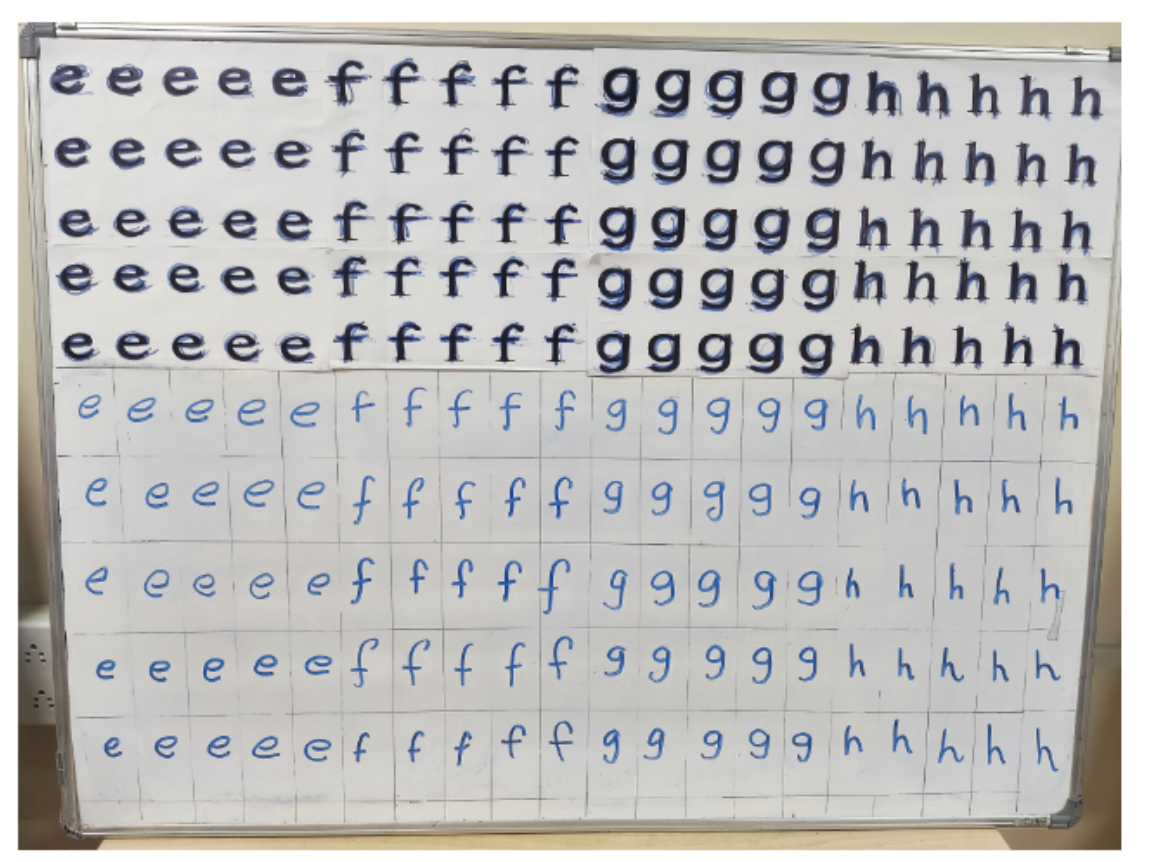} 
\caption{An illustration of the setup build in the ubiquitous computing lab for dataset collection.}
\label{fig4}
 \end{figure}

By addressing these challenges and considering the specific requirements of collecting sensory data using IMU sensors on a whiteboard, you can set up an effective data collection process to gather valuable insights for your research or application. For effectively and paralley performing  the data collection inside our ubiquitous computing lab, we have used four white boards. Each student volunteers write alphabets using the developed data collection device mounted over the white board markers. We have draw some initial letters on the board so that the volunteer can easily practice out the character writing prior to the actual writing on the board. Fig.\ref{fig4} illustrates the data collection steup at the ubiquitous computing lab IIT (BHU) Varanasi, India.

\subsection{Experimental evidences 2}
\begin{figure}[h]
 \centering
\includegraphics[scale=0.52]{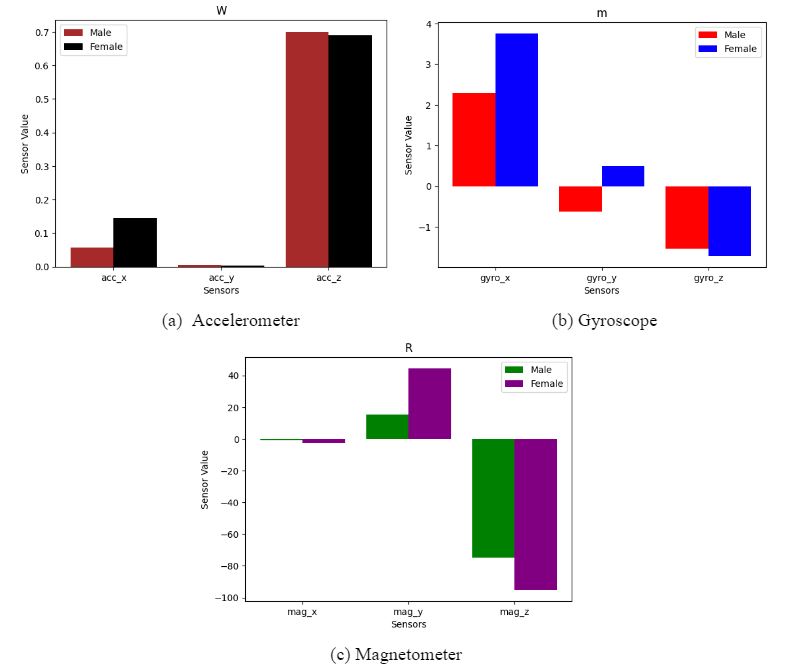} 
\caption{Illustration of impact of male vs female volunteers on the values of the accelerometer for character W, gyroscope for the character m, and magnetometer for the character R.}
\label{fig5}
 \end{figure}

We present some experimental evidence, which are gathered through controlled experiments or observations to test the collected dataset.  Fig.~\ref{fig5} provide a visual representation, depicting the impact of male and femal volunteers on the value of accelerometer, gyroscope, and magnetometer sensors on selected characters (W, m, and R).

\begin{figure}[h]
 \centering
\includegraphics[scale=0.32]{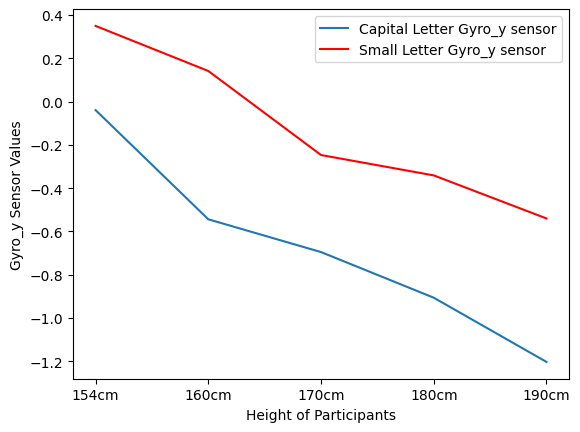} 
\caption{Illustration of the impact of volunteers height on the angle of writing, given whiteboard is at the same height for all volunteers.}
\label{fig6}
 \end{figure}

From Fig.~\ref{fig6}, we can make following important observations, where, gyro\_y represents the y-axis of the gyroscope sensor.

\subsubsection{Height and Writing Angle} When individuals write on a whiteboard, the angle at which they hold their hand and write can vary depending on their height. Taller individuals tend to have their hands positioned at a higher level compared to shorter individuals when writing on the whiteboard. This is because the height difference affects the natural positioning of their arm and hand while reaching the whiteboard surface. 

\subsubsection{Gyro\_y Sensor} The gyro\_y sensor is a type of sensor that measures the rotation or angular velocity around the y-axis. It can be used to detect changes in the angle or orientation of an object. Thus, we observe that the gyro\_y sensor values decrease as height increases. This suggests that the sensor is capturing a lower angular velocity around the y-axis for taller individuals compared to shorter individuals while they are writing on the whiteboard. In other words, taller individuals tend to have a smaller rotational movement or a flatter angle while writing characters, resulting in lower values recorded by the gyro\_y sensor.

\section{Volunteers for data collection}\label{vol}
This section provides a detailed description of the student participants who volunteered to contribute to the collection of a large-size sensory dataset using an IMU sensor. We begin by discussing the various challenges encountered during the selection process of the volunteers. Following that, we present a concise overview of the attributes that may have an impact on the dataset collection, such as height and hand orientation. Furthermore, this section includes a table or pie chart depicting the distribution of participants' height, weight, and age. Additionally, we discuss the geographical diversification of student volunteers by incorporating an Indian map to illustrate their regional representation.

The fundamental challenges encountered while selecting the volunteers for data collection are discussed as the following questions:
\begin{itemize}

\item[\textbf{Q1:}] How did you ensure consistent sensor placement across participants of varying heights?

\item[\textbf{Q2:}] How did you address the potential impact of height differences on sensor alignment? 

\item[\textbf{Q3:}] How to ensure  that the selected volunteers represent the target population or the group being studied is essential for obtaining accurate and reliable data? 

\item[\textbf{Q4:}] How to achieve a diverse and representative sample as it be can be challenging due to factors such as demographics, availability, and willingness to participate.

\item[\textbf{Q5:}] How to handle biasness that can occur when the selection process favours certain types of individuals over others, leading to a skewed representation of the volunteers?

\item[\textbf{Q6:}] How to tackle the sampling errors that can occur due to random variability or issues with the sampling method, resulting in a sample that does not accurately reflect the volunteers' characteristics?

\item[\textbf{Q7:}] Importantingly how to find and recruit volunteers who are willing to participate in the data collection process?

\item[\textbf{Q8:}] How to achieve a diverse and representative sample as it be can be challenging due to factors such as demographics, availability, and willingness to participate.

\item[\textbf{Q9:}] How to provide a favorable and noise-free data collection environment to reduce the hesitation or reluctance of the volunteers during the data collection?

\item[\textbf{Q10:}] How to handle ethical standards in volunteer selection?

\item[\textbf{Q11:}] What are the steps required to obtain informed consent, protect participant privacy and confidentiality, and address any potential risks or harms associated with the study?

\item[\textbf{Q12:}] How do handling constraints such as time, budget, and personnel can limit the number of volunteers that can be selected or the methods used for recruitment and data collection?

\end{itemize}    

\begin{figure}[h]
 \centering
\includegraphics[scale=0.18]{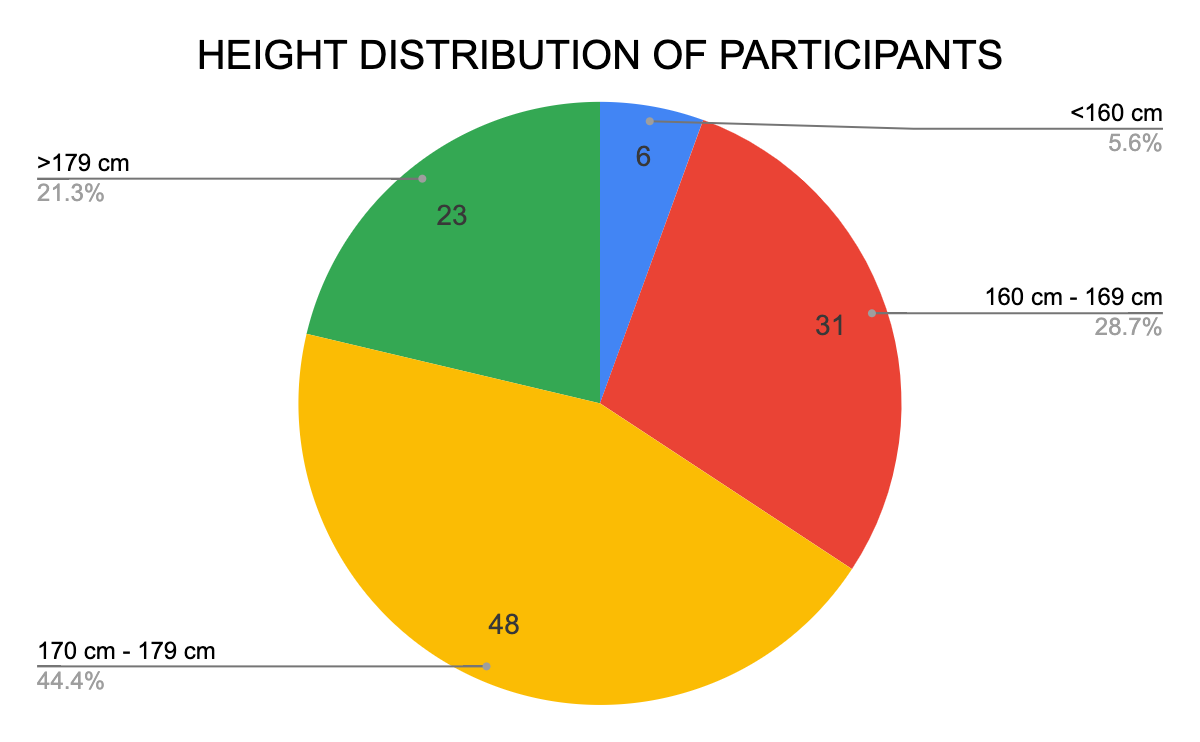} 
\caption{Illustration of height distribution of student participants as volunteers for collecting the handwritten English alphabet characters dataset using IMU sensor.}
\label{fig7}
 \end{figure}

\begin{figure*}[h]
 \centering
\includegraphics[scale=0.54]{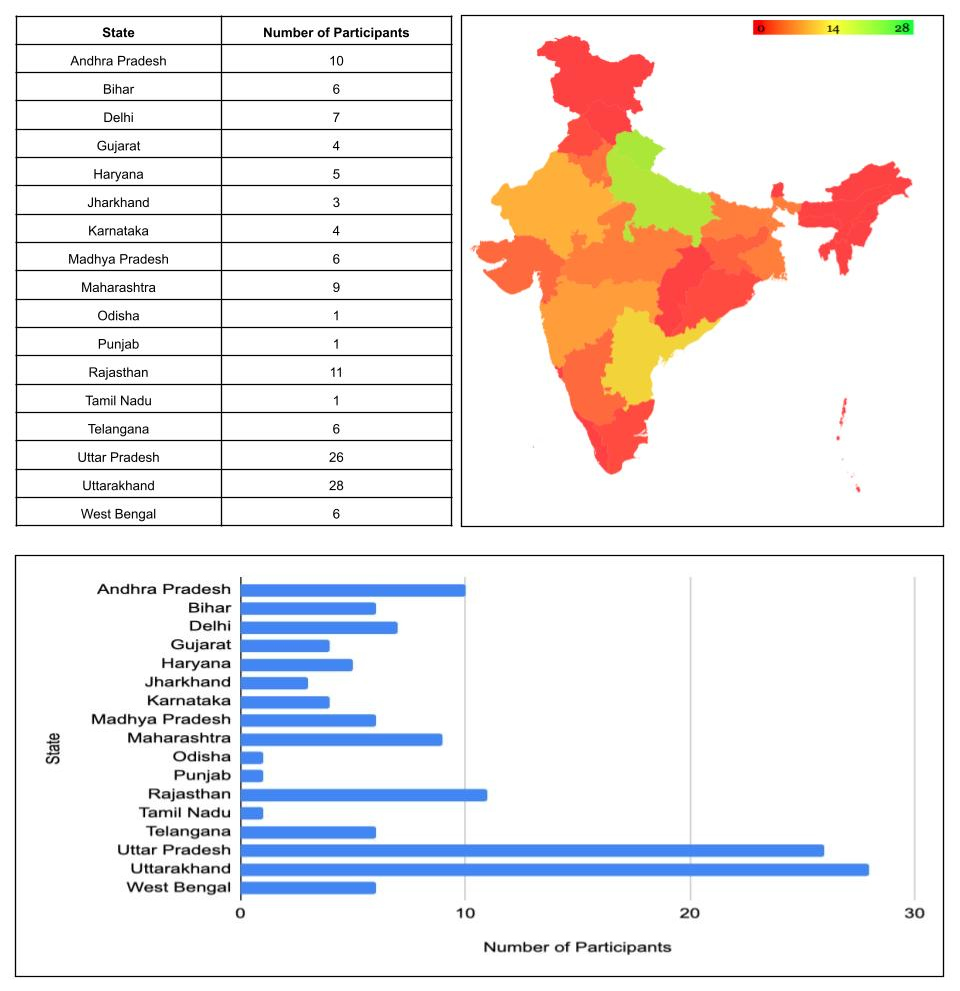} 
\caption{Illustration of volunteer distribution in terms of geographic regions in India.}
\label{fig8}
 \end{figure*}

During the data collection phase, we carefully selected 124 student volunteers to participate in the study. Their task was to write the English alphabet, both in capital and small letters, on a whiteboard. Each student was given specific instructions to write each character 50 times during various shifts over a period of two months. The student volunteers varied in terms of their heights and genders, which introduced diversity into the dataset, as shown in Fig.~\ref{fig7}. Moreover, the student volunteers are from different locations in India (as in Fig.~\ref{fig8}), making a high level of heterogeneity in terms of culture and behaviour.

 Additionally, each individual had their own unique writing pattern and speed, adding further variability to the collected data. By allowing the volunteers to write the characters over a two-month period, we aimed to minimize biases that could arise from monotonous repetition. The aforementioned characteristics of the dataset result in a high level of heterogeneity. This diversity makes the collected dataset more suitable for analyzing and training different models in order to simulate and address realistic scenarios. 

The presence of different writing styles, speeds, and durations in the dataset reflects the real-world variations that exist in individuals' writing behaviours. By incorporating this heterogeneity into the dataset, we can obtain a more comprehensive understanding of the challenges and variations that may be encountered in practical applications. This diverse dataset enables the development and testing of models that can handle a wider range of writing styles and conditions, ultimately leading to more robust and adaptable solutions in the field of character recognition or other related tasks.
 
\section{Dataset description}\label{des}
The data collection process involved the participation of 124 students who were assigned the task of writing 52 letters of the English alphabet. This included 26 lowercase letters (a-z) and 26 uppercase letters (A-Z). The experiment was conducted over a period of two months, ensuring a substantial amount of data was gathered. Each student visited the laboratory six times a week to perform the writing task. During each session, they wrote two characters on the whiteboard. The students were instructed to write each character 50 times on the board. To capture the timing and duration of each writing instance, a button was provided to the students. They would press the button before starting to write a character and again immediately after completing the character.

\begin{figure}[h]
 \centering
\includegraphics[scale=0.27]{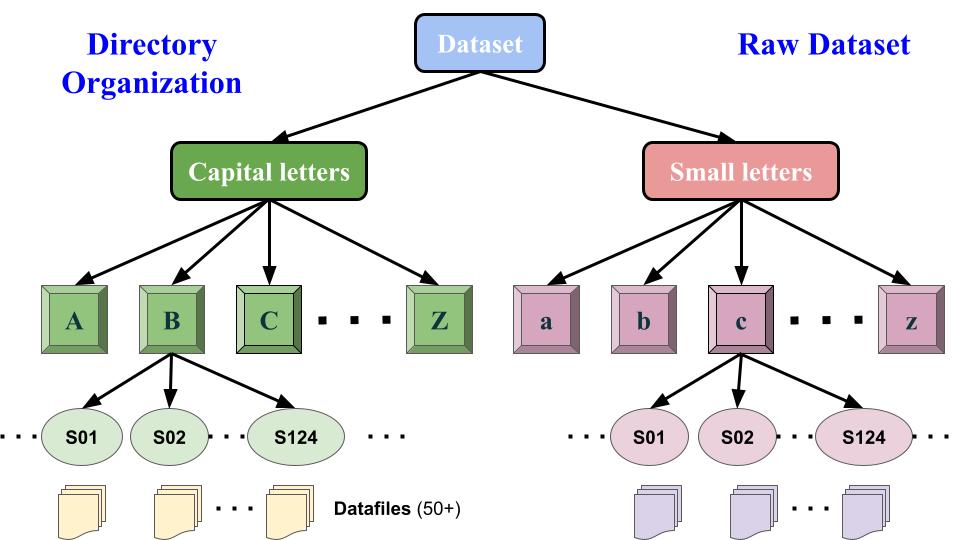} 
\caption{An illustration of directory organization in the collected alphabet dataset.}
\label{fig9}
 \end{figure}

\begin{figure}[h]
 \centering
\includegraphics[scale=0.16]{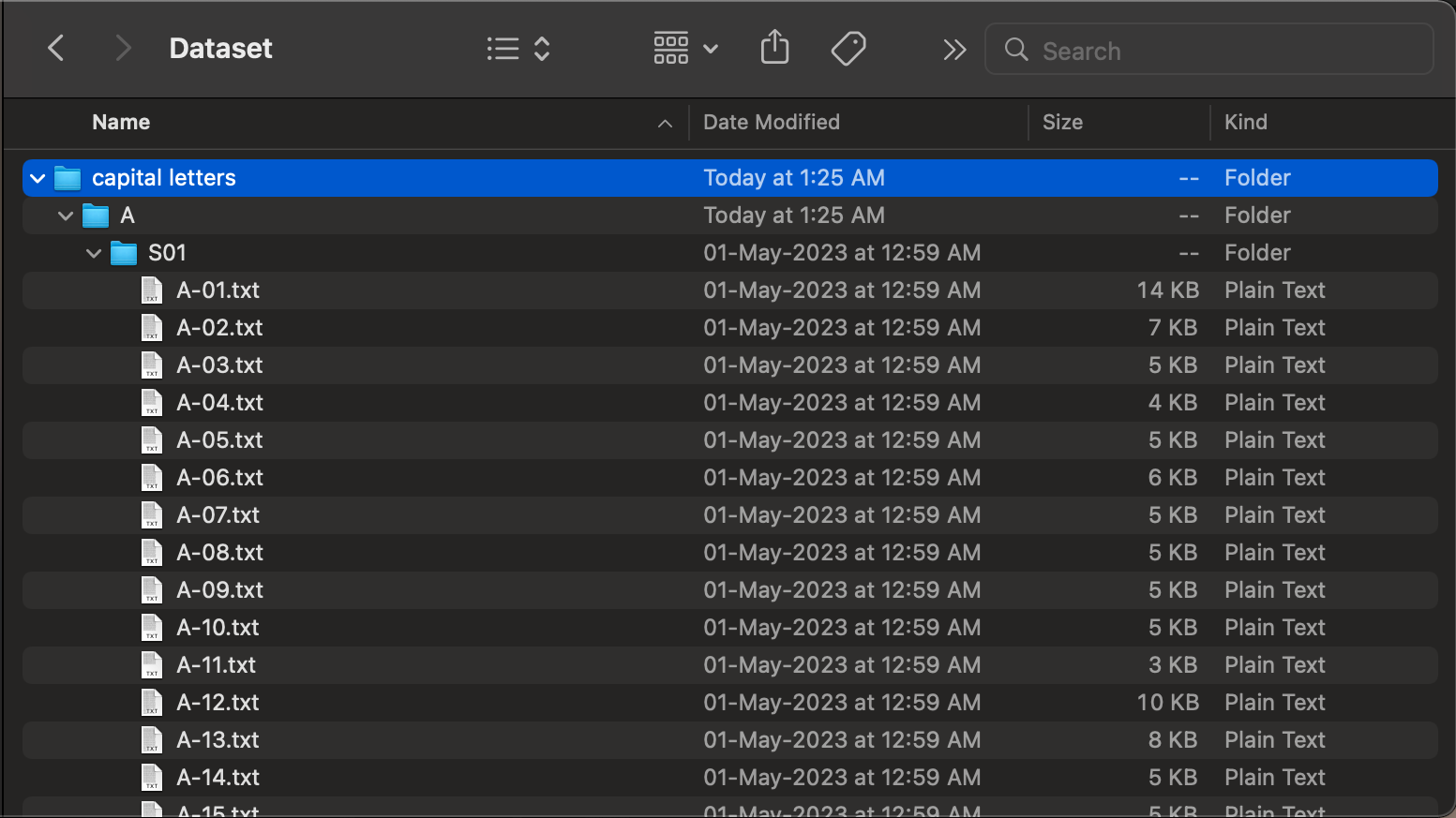} 
\caption{An illustration of the internal directory (S01), which contains different text files for letter A.}
\label{fig10}
 \end{figure}

During the writing process, a sensor was employed to record data. When the student was actively writing a character, the sensor registered a value of 1, along with other sensory measurements. On the other hand, when the student was not writing a character, the sensor recorded a value of 0, along with the corresponding sensory data. This allowed for the differentiation between active writing periods and non-writing periods. Throughout the dataset collection, the sensors recorded data from all 50 instances of each character being written by the students. This comprehensive data collection approach enabled the capture of multiple repetitions of each character, providing a rich dataset for analysis and modelling purposes. By combining information on the sensory values, timing, and the distinction between writing and non-writing periods, the collected data offers valuable insights into the writing behaviour and patterns exhibited by the students. The sensory dataset is stored in \textit{csv} files, following the directory hierarchy as shown in Fig.\ref{fig9}. Furthermore, Fig.~\ref{fig10} illustrates the inner directory organization of the dataset.

 During the data collection process, a specific protocol was followed to organize and record the data for each participant. Each file in the dataset represents a particular character, and within that file, the participant wrote the same character 50 times. All 50 instances of writing that character by a participant were recorded in this file. To initiate the recording of each writing instance, the student pressed a button. This action signalled the start of writing and corresponded to a value of 1 in the dataset. Subsequently, each time the participant completed writing the character, they pressed the button again, resulting in a value of 0 being recorded in the dataset. Along with these binary values, readings from sensors such as accelerometers, gyroscopes, and magnetometers were also captured.

The process was repeated for every student. For each participant, 50 different files were created, each representing the same character they wrote, but at different instances. This ensured that there was a separate file for every instance of writing the character by each student. By organizing the data in this manner, researchers were able to maintain a systematic record of each participant's writing instances. This allowed for further analysis and examination of patterns, variations, and trends within and across participants. Additionally, having individual files for each instance and participant facilitated the isolation and study of specific writing behaviours, helping to identify unique characteristics or challenges associated with writing a particular character. Overall, this approach ensured that the data collected was structured and categorized in a way that facilitated subsequent analysis and modelling tasks. The number of data instances against each character is graphically shown at the bottom of this document.

\subsection{Dataset preprocessing}
We shortlist all files ranging from 40 to 80 lines of sensory data per instance. We add padding to all the files at the top and bottom to make it 80 lines. Each instance of a character has 80 lines of sensory data. The dataset is partitioned into two sets, where 80\% of the students were selected for generating the training data and 20\% for the test data. In the case of sensor overflow values, we replace them with similar sensor axis data from the dataset. The preprocessing steps are discussed as follows:

\begin{enumerate}
\item \textit{Shortlisting Files:} The files containing sensory data are examined, and those ranging from 40 to 80 lines are selected for further processing. This step ensures that only files within this specific range of lines are considered for analysis. 

\item \textit{Padding for Standardization:} To maintain consistency in the dataset, padding is applied to the selected files. Padding involves adding extra lines of sensory data at the top and bottom of each file to make it a standardized length of 80 lines. By doing so, all instances of characters in the dataset have the same number of lines. 

\item \textit{80 Lines of Sensory Data per Instance:} After the padding is applied, each instance of a character now consists of 80 lines of sensory data. This standardized length allows for easier comparison and analysis across different instances and characters. 

\item \textit{Dataset Partitioning:} The dataset is then divided into two separate sets: the training data and the test data. Approximately 80\% of the students' data is allocated to the training set, which will be used for generating models and training them. The remaining 20\% of students' data is assigned to the test set, which will be used to evaluate the performance of the trained models. 

\item \textit{Handling Sensor Overflow Values:} In some cases, the sensory data may contain overflow values, which are values that exceed the expected range or capacity of the sensor. To address this issue, the overflow values are replaced with similar sensor axis data from within the dataset. This ensures that the dataset remains consistent and coherent, even in the presence of overflow values, by substituting them with appropriate data points. 

\end{enumerate}

By following these steps, the dataset is processed, standardized, and divided into training and test sets. These preparations facilitate the training of models on the training data and the subsequent evaluation of their performance using the test data. 

To refine the dataset, we implemented a selection process for the files. We specifically shortlisted files that fell within the file size range of 4 to 40 kb. Additionally, we conducted tests on files of various length ranges, including 3 to 30 kb, 3 to 50 kb, 5 to 10 kb, and 5 to 20 kb. Any files that did not fall within these specified ranges were removed from consideration. Once the files were selected, we performed an analysis on the number of lines present in each file with the same instance length. We calculated the average number of lines across these files. For files that had fewer lines than the average, we added additional lines filled with zeros at the top and bottom to match the average line count. On the other hand, for files that had more lines than the average, we removed half of the lines from the top and half from the end to bring them in line with the average. Furthermore, we enforced a requirement that the character being written must be present in the first line of each instance. If this condition was not met, the file was removed from the dataset. To ensure data consistency, we eliminated any occurrences of "nan" (indicating missing data) or "ovf" (indicating an overflow) in the sensor readings. If a sensor reading contained one of these values and was present in the file, it was replaced with a zero to maintain uniformity in the dataset. By applying these selection criteria and data-cleaning techniques, we aimed to create a refined dataset that would be more suitable for analysis, modelling, and training purposes.

For additional experimental analysis, we performed the following actions:

\begin{itemize}

\item \textit{Addition of Noisy Samples:} We augmented the dataset by adding 50,000 samples with a 3\% noise level. This means that 3\% of the characters in each sample were randomly altered or distorted to simulate real-world scenarios where noise or errors may be present. The noise was introduced to increase the variability and robustness of the dataset. 

\item \textit{Standard Deviation (Std Dev) of each char-median value:} We calculated the standard deviation of the median value for each character in the dataset. This measurement helps to quantify the variability or spread of the median values across different instances of the same character. The std dev values provide insights into the consistency or variation of character representations within the dataset. 

\item \textit{Range of Values:} 
\begin{itemize}
        \item \textit{a-range:} The range for the character "a" was determined to be 70 data points. This implies that the values associated with the character "a" span a range of 70 data points in a line.\\
        \item \textit{b-range:} The range for character "b" was found to be 80 data points, indicating that the values associated with the character "b" cover a range of 80 data points in a line.
\end{itemize}
\end{itemize}

\subsection{Data distribution of upper-case letters}
\begin{figure}[h]
\centering
\includegraphics[scale=0.55]{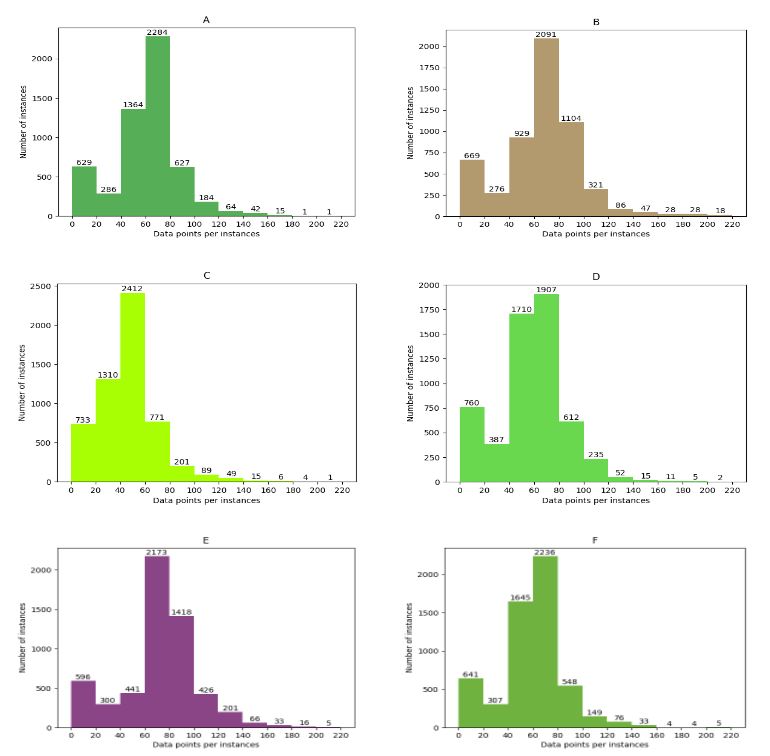}
\caption{Data distribution of upper-case letters in the collected dataset.}
\label{fig11}
 \end{figure}

Fig.~\ref{fig11} depicts the data distribution of upper-case letters in the collected dataset.

\subsection{Data distribution of lower-case letters}
\begin{figure}[h]
\centering
\includegraphics[scale=0.55]{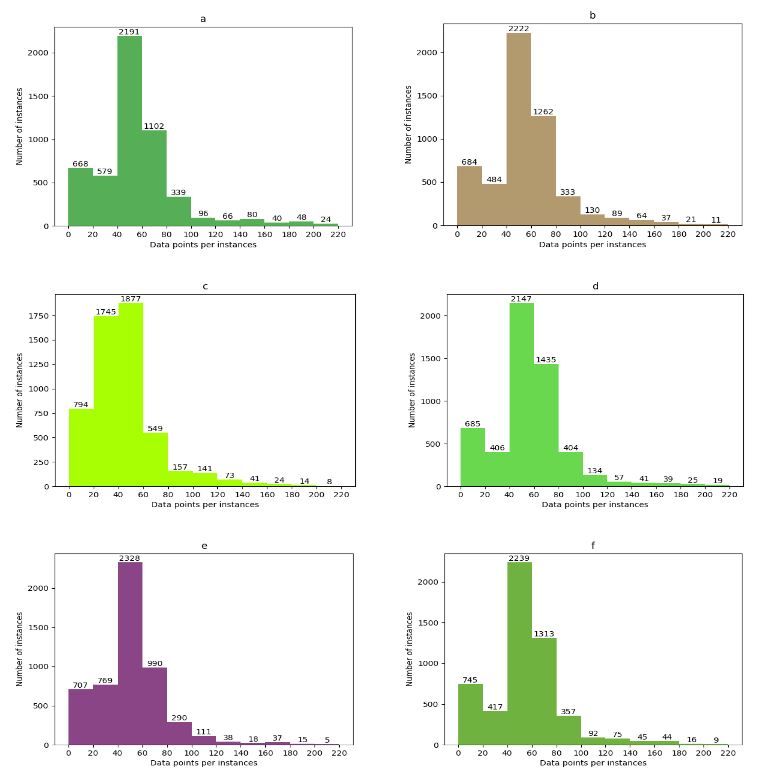}
\caption{Data distribution of lower-case letters in the collected dataset.}
\label{fig12}
 \end{figure}

Fig.~\ref{fig12} depicts the data distribution of lower-case letters in the collected dataset.

\section{Conclusion}\label{conc} 
The collected dataset for handwritten English alphabets recognition using IMU showcases the potential of leveraging IMU technology to accurately recognize and classify handwritten alphabets. The findings highlight the advantages of capturing dynamic hand movements and orientation through IMUs, leading to improved accuracy and potential applications in education, digital platforms, and human-computer interaction. Future research can focus on refining algorithms, exploring integration with other modalities, and assessing performance in real-world scenarios. Overall, IMUs offer innovative possibilities for transforming written word interactions and advancing various fields.

\bibliographystyle{IEEEtran}
\bibliography{lpencil}

\end{document}